\documentclass[journal]{IEEEtran}
\usepackage{times,amsmath,epsfig}
\usepackage{graphicx}
\usepackage{subfigure}
\usepackage{amsmath}
\usepackage{color}
\usepackage{comment}
\usepackage{mathtools}
\usepackage{algorithm}
\usepackage{algorithmic}
\usepackage{amsfonts,amssymb}
\usepackage{theorem}

\newtheorem{lemma}{Lemma}

\renewcommand{\algorithmicrequire}{\textbf{Input:}}
\renewcommand{\algorithmicensure}{\textbf{Output:}}

\usepackage{hyperref}
\usepackage{color}
\usepackage[normalem]{ulem}

\newcounter{ToDo}
\newcounter{gaocomm}
\newcounter{Note}
\definecolor{blue-violet}{rgb}{0.54, 0.17, 0.89}
\definecolor{mygreen}{rgb}{0.0, 0.5, 0.0}
\definecolor{awesome}{rgb}{1.0, 0.13, 0.32}
\definecolor{bostonuniversityred}{rgb}{0.8, 0.0, 0.0}

\begin{document}
\title{\textbf{Tensorial Recurrent Neural Networks for Longitudinal Data Analysis}}
\author{Mingyuan Bai, 
        Boyan Zhang and Junbin Gao 
\IEEEcompsocitemizethanks{
\IEEEcompsocthanksitem Mingyuan Bai and Junbin Gao 
are with with the Discipline of Business Analytics, The University of Sydney Business School, The University of Sydney, NSW 2006, Australia. \protect E-mail: mbai8854@uni.sydney.edu.au; junbin.gao@sydney.edu.au
\IEEEcompsocthanksitem Boyang Zhang is  School of Information Technologies,
The University of Sydney, NSW 2006, Australia. \protect E-mail: bzha8220@uni.sydney.edu.au 
}
}
\maketitle
\begin{abstract}
Traditional Recurrent Neural Networks assume vectorized data as inputs. However many data from modern science and technology come in certain structures such as tensorial time series data.  To apply the recurrent neural networks for this type of data, a vectorisation process is necessary, while such a vectorisation leads to the loss of the precise information of the spatial or longitudinal dimensions. In addition, such a vectorized data is not an optimum solution for learning the representation for the longitudinal data. 
In this paper, we propose a new variant of tensorial neural networks which directly take tensorial time series data as inputs. 
We call this new variant as Tensorial Recurrent Neural Network (TRNN). The proposed TRNN is based on tensor Tucker decomposition.
\end{abstract}

\section{Introduction}\label{Sec:1}

In recent years, the interests in time series with sequential effects among the data have been constantly growing, in both academic field and industry.  These interests are from the development of technology and social science including but not limited to, multimedia, social network and economic and political network, especially international relationship study. 

Time series data acquired from many discipline is not only in large volume  in terms of time, but also in more ever complicated structures, such as in multi- and high-dimension.  The rise of massive multi-dimensional data has led to new demands for Machine Learning (ML) systems to learn complex models with millions to billions of parameters for new types of data structures, that promise adequate capacity to digest massive datasets and offer powerful predictive analytics thereupon. As tensorial data come with a special spatial structure, it is highly desired to maintain this structure information in learning process. We have seen the most recent development in deep learning architecture for multi-dimensional tensor data, extending conventional (vector) neural networks to structured data, such as the matrix neural network \cite{GaoGuoWang2017,IonescuVantzosSminchisescu2015a}, two independent works on tensorial neural networks \cite{BaiZhangGao2017,ChienBao2017}, graph data \cite{LiTarlowBrockschmidtZemel2016,SeoDefferrardVandergheynstBresson2017}, and even neural networks for manifold-valued data \cite{HuangGool2017}.

The recurrent neural networks (RNN) as a commonly applied tool in longitudinal data analysis have constantly been investigated in the last couple of decades with many successful applications, such as language processing \cite{MulderBethardMoens2015}, speech recognition \cite{GravesMohamedHinton2013}, and human action recognition \cite{LiuShahroudyXuWang2016,ZhangLanXingZengXueZheng2017} etc. There are many different architectures for RNN such as the basic recurrent network,   Elman networks or Jordan networks,  long short-term memory (LSTM), and gated recurrent unit (GRU) etc.  

The most recent developments of recurrent neural networks are generally focused on the LSTM model. For example, the LSTM has been combined with the convolutional neural networks (CNN) for sequence representation learning \cite{GanyPuyHenaoyLiyHezCarin2016}. However the traditional LSTM model can only deal with vectorised data, which leads to the loss of some spatial information for multidimensional time series. 

Our intention in this paper is to propose a fully tensorial connected neural networks for tensorial longitudinal data. A recent paper has also considered tensorial structure in the classical recurrent neural networks, however the main purpose was to reduce the number of networks parameters when vectorial data are in very high-dimension, \cite{JoseCisseFleuret2017}.

The rest of this paper is organized as follows. Section \ref{Sec:2} introduces basic recurrent neural network and two types of tensorial RNN, i.e., tensorial LSTM (tLSTM) and tensorial GRU (tGRU).   In Section \ref{Sec:3}, we derive the backpropagation algorithms for the proposed tLSTM and rGRU.    In Section \ref{Sec:4}, experimental results are presented to evaluate the performance of the proposed models. Finally, conclusions and future works are summarized in Section \ref{Sec:5}.

\section{Tensorial Recurrent Neural Networks} \label{Sec:2} 
The simple building block for RNN is expressed  in the following
forward mapping, Elman model \cite{Elman1990},
\begin{align}
\begin{aligned}
\mathbf h_t =& \sigma_h (\mathbf W_{hx}\mathbf x_t + \mathbf W_{hh}\mathbf h_{t-1} + \mathbf b_h)
\\
\mathbf y_t =&\sigma_y(\mathbf W_{hy}\mathbf h_t + \mathbf b_y)
\end{aligned} \label{Eq1}
\end{align}
or in Jordan model \cite{Jordan1997},
\begin{align}
\begin{aligned}
\mathbf h_t =& \sigma_h (\mathbf W_{hx}\mathbf x_t + \mathbf W_{hh}\mathbf y_{t-1} + \mathbf b_h)
\\
\mathbf y_t =& \sigma_y(\mathbf W_{hy}\mathbf h_t + \mathbf b_y)
\end{aligned} \label{Eq2}
\end{align}
Jordan model further passes on the output information at time $t$ to the next time $t+1$ as inputs.

In this note, we will focus on Elman model \eqref{Eq1}. However we will consider the setting for tensorial longitudinal data, in general, denoted by
\[
\mathcal{D} = \{(\mathcal{X}_t, \mathcal{Y}_t)\}^T_{t=1}.
\]
where each independent data $\mathcal{X}_t$ is a tensor of $D$-ways (the tensor dimension) and the response data $\mathcal{Y}_t$ could be a scalar, a vector or more general a tensor of the same dimension as $\mathcal{X}_t$.

The two most popular recurrent neural network architectures are the Long Short-Term Memory Units (LSTMs) and the Gated Recurrent Units (GRUs), which will be extended for tensors data.

\subsection{Tensorial LSTM (tLSTM)}
LSTMs were introduced in \cite{HochreiterSchmidhuber1997} and further modernized by   many people, e.g. \cite{GersSchmidhuberCummins2000}.  The application has demonstrates that LSTMs work tremendously well on a large variety of problems, and are now widely used.

Based on the classic LSTMs, we propose the following tensorial LSTM,  

\begin{align}
\mathcal{F}_t =& \sigma_g(\mathcal{H}_{t-1}\times_1 \mathbf W_{f1}\times_2 \mathbf W_{f2}\times \cdots \times_D\mathbf{W}_{fD}  \notag\\
&\phantom{\sigma_g(} + \mathcal{X}_{t}\times_1 \mathbf U_{f1}\times_2 \mathbf U_{f2}\times \cdots \times_D\mathbf{U}_{fD} + \mathcal{B}_f) \label{Eq3}\\
\mathcal{I}_t =& \sigma_g(\mathcal{H}_{t-1}\times_1 \mathbf W_{i1}\times_2 \mathbf W_{i2}\times \cdots \times_D\mathbf{W}_{iD}  \notag\\
&\phantom{\sigma_g(} + \mathcal{X}_{t}\times_1 \mathbf U_{i1}\times_2 \mathbf U_{i2}\times \cdots \times_D\mathbf{U}_{iD} + \mathcal{B}_i) \label{Eq4}\\
\mathcal{O}_t =& \sigma_g(\mathcal{H}_{t-1}\times_1 \mathbf W_{o1}\times_2 \mathbf W_{o2}\times \cdots \times_D\mathbf{W}_{oD}  \notag\\
&\phantom{\sigma_g(} + \mathcal{X}_{t}\times_1 \mathbf U_{o1}\times_2 \mathbf U_{o2}\times \cdots \times_D\mathbf{U}_{oD} + \mathcal{B}_o) \label{Eq5}\\
\widehat{\mathcal{C}}_t =& \sigma_c(\mathcal{H}_{t-1}\times_1 \mathbf W_{c1}\times_2 \mathbf W_{c2}\times \cdots \times_D\mathbf{W}_{cD}  \notag\\
&\phantom{\sigma_g(} + \mathcal{X}_{t}\times_1 \mathbf U_{c1}\times_2 \mathbf U_{c2}\times \cdots \times_D\mathbf{U}_{cD} + \mathcal{B}_c) \label{Eq6}\\
\mathcal{C}_t =& \mathcal{F}_t\circ \mathcal{C}_{t-1} + \mathcal{I}_t \circ \widehat{\mathcal{C}}_t \label{Eq7}\\
\mathcal{H}_t =& \mathcal{O}_t \circ \sigma_h(\mathcal{C}_t) \label{Eq8}
\end{align}
where the operator $\circ$ denotes the Hadamard product, i.e., the entry-wise product, and $\mathbf W_{\cdot d}$ as well as $\mathbf W_{\cdot d}$ are matrices in relevant order, applied on hidden tensorial variables and input tensorial variables in terms of tensorial mode product \cite{KoldaBader2009}, and all $\mathcal{B}_{\cdot}$ are tensorial biases.

Note $\mathcal{O}_t$ is not the actual output of the LSTM. In fact, $\mathcal{O}$ will be jointly regulated by both $\mathcal{I}_t$ and $\mathcal{C}_t$ to make the potential output from the hidden variable $\mathcal{H}_t$ as in \eqref{Eq8}. Depending on the type of response data $\mathcal{Y}_t$, we may apply an extra layer of neural network on the top of $\mathcal{H}_t$ to convert the tensor hidden $\mathcal{H}_t$ to the shape/structure of $\mathcal{Y}_t$. For the sake of notation simplicity, we assume the transformed output is denoted by $\widehat{\mathcal{O}}_t$. 

\subsection{Tensorial GRU (tGRU)}
The Gated Recurrent Unit (GRU) was introduced in \cite{ChoMerrieenboerGuelccehreBahdanauBougaresSchwenkBengio2014} with a slightly more dramatic variation on the LSTM.  Similar to GRU, in our proposed tensorial GRU, the forget $\mathcal{F}_t$ and the input $\mathcal{I}_t$ gates are to be combined into a single ``update gate''.  tGRU is simpler than the aforementioned tLSTM models.  

\begin{align}
\mathcal{R}_t =& \sigma_g(\mathcal{H}_{t-1}\times_1 \mathbf W_{r1}\times_2 \mathbf W_{r2}\times \cdots \times_D\mathbf{W}_{rD}  \notag\\
&\phantom{\sigma_g(} + \mathcal{X}_{t}\times_1 \mathbf U_{r1}\times_2 \mathbf U_{r2}\times \cdots \times_D\mathbf{U}_{rD} + \mathcal{B}_r) \label{Eq9}\\
\mathcal{Z}_t =& \sigma_g(\mathcal{H}_{t-1}\times_1 \mathbf W_{z1}\times_2 \mathbf W_{z2}\times \cdots \times_D\mathbf{W}_{zD}  \notag\\
&\phantom{\sigma_g(} + \mathcal{X}_{t}\times_1 \mathbf U_{z1}\times_2 \mathbf U_{z2}\times \cdots \times_D\mathbf{U}_{zD} + \mathcal{B}_z) \label{Eq10}\\
\widehat{\mathcal{R}}_t =& \mathcal{R}_t \circ \mathcal{H}_{t-1} \label{Eq11}\\
\widehat{\mathcal{H}}_t =& \sigma_h(\widehat{\mathcal{R}}_{t}\times_1 \mathbf W_{h1}\times_2 \mathbf W_{h2}\times \cdots \times_D\mathbf{W}_{hD}  \notag\\
&\phantom{\sigma_g(} + \mathcal{X}_{t}\times_1 \mathbf U_{h1}\times_2 \mathbf U_{h2}\times \cdots \times_D\mathbf{U}_{hD} + \mathcal{B}_h) \label{Eq12}\\
\mathcal{H}_t =& \mathcal{Z}_t \circ \mathcal{H}_{t-1} + (1 - \mathcal{Z}_t) \circ \widehat{\mathcal{H}}_t. \label{Eq13}
\end{align}
%
%

Similar to tLSTM, we will add an additional transform mapping the hidden variables $\mathcal{H}_t$ to match the response variable $\mathcal{Y}_t$.

\section{Recurrent BP Algorithm} \label{Sec:3} 
\subsection{Loss Function}
According to the way how data is presented, we propose three types of loss functions.

\textit{Loss function for single data series.} The training data is presented as a single time series and we will apply the LTSM or GRU on the series, and collect their outputs at each time point. The simple loss at each time is defined as
\begin{align*}
\ell_s(t) = l(\mathcal{Y}_t, \widehat{\mathcal{O}}_t) 
\end{align*}
which is the building block for all the other overall loss function.
Please note that $\widehat{\mathcal{O}}_t$ is calculated through the recurrent networks from the input $\{\mathcal{X}_1, \mathcal{X}_2, ..., \mathcal{X}_t\}$.  $l$ is a loss function such as the usual squared loss function for regression or the cross-entropy loss for classification.

The overall loss is defined as
\begin{align}
\ell_s = \sum^T_{t=1}\ell_s(t) =\sum^T_{t=1}l(\mathcal{Y}_t, \widehat{\mathcal{O}}_t)  \label{Eq14}
\end{align}

\textit{Loss function for multiple data series with same length/duration}.  Most of time, we will use a recurrent network structure with a certain duration. In this case, we will assume that the training data consist of a number of training series,
\[
\mathcal{D} = \{(\mathcal{X}_{j1}, ..., \mathcal{X}_{jT}), (\mathcal{Y}_{j})\}^N_{j=1}
\]
The loss for the $j$-th case is only calculated at time $T$ as
\[
\ell_T(j) = l(\mathcal{Y}_j, \widehat{\mathcal{O}}_{jT})
\]
where $\widehat{\mathcal{O}}_{jT}$ is the last output of LSTM or GRU from the input series $(\mathcal{X}_{j1}, ..., \mathcal{X}_{jT})$.
Hence the overall loss is
\begin{align}
\ell_T = \sum^N_{j=1}\ell_T (j) =\sum^N_{j=1}l(\mathcal{Y}_j, \widehat{\mathcal{O}}_{jT}). \label{Eq15}
\end{align}

Similar to the simple series case, if the response is a series $(\mathcal{Y}_{j1}, ..., \mathcal{Y}_{jT})$, then the loss can be revised as
\begin{align}
\ell_m = \sum^N_{j=1}\sum^T_{t=1}\ell_m(t,j) =\sum^N_{j=1}\sum^T_{t=1}l(\mathcal{Y}_{jt}, \widehat{\mathcal{O}}_{jt}).\label{Eq16}
\end{align}

\textit{Loss function for Panel Data} In many application case particularly for panel data, the duration or period $T$ for each series $(\mathcal{X}_{j1}, ..., \mathcal{X}_{jT})$ may be different, thus the recurrent network will run through different loops. Suppose the data are
\[
\mathcal{D} = \{(\mathcal{X}_{j1}, ..., \mathcal{X}_{jT_j}), (\mathcal{Y}_{j})\}^N_{j=1},
\]
then the loss can be defined as
\begin{align}
\ell_{Tp} = \sum^N_{j=1}\ell_{Tp}(j) =\sum^N_{j=1}l(\mathcal{Y}_j, \widehat{\mathcal{O}}_{jT_j}).\label{Eq17}
\end{align}
or
\begin{align}
\ell_{mp} = \sum^N_{j=1}\sum^{T_j}_{t=1}\ell_m(t,j) =\sum^N_{j=1}\sum^{T_j}_{t=1}l(\mathcal{Y}_{jt}, \widehat{\mathcal{O}}_{jt}). \label{Eq18}
\end{align}

Loss function \eqref{Eq14} is a special case of loss function \eqref{Eq16} when $N=1$. For the BP algorithm in the next subsection, we will focus on losses \eqref{Eq15} and \eqref{Eq16}. Similarly losses \eqref{Eq17} and \eqref{Eq18} can be processed in the similar way as for \eqref{Eq15} and \eqref{Eq16}, respectively.
\subsection{BP Algorithm}

The major difference between the proposed tensorial RNN (tRNN) and the (vectorial) RNN is that the vectorial linear mapping has been replaced with the tensor multiple linear mapping, i.e., Tucker multiplication, \cite{KoldaBader2009}. Let us denote Tucker mapping by
\begin{align*}
\mathcal{M}^{\alpha}_t =&\mathcal{H}_{t-1}\times_1 \mathbf W_{\alpha 1}\times_2 \mathbf W_{\alpha 2}\times \cdots \times_D\mathbf{W}_{\alpha D}  \notag\\
& + \mathcal{X}_{t}\times_1 \mathbf U_{\alpha 1}\times_2 \mathbf U_{\alpha 2}\times \cdots \times_D\mathbf{U}_{\alpha D} + \mathcal{B}_{\alpha}  
\end{align*}
where $\alpha$ can be either $f$, $i$, $o$, $c$, $r$, $z$ or $h$. When $\alpha = h$, we replace $\mathcal{H}_{t-1}$ with $\widehat{\mathcal{R}}_t$ in \eqref{Eq12}.

First we introduce the results from \cite{Kolda2006} without proof.
\begin{lemma} Denote by $|\mathcal{M}|$ and $|\mathcal{H}|$ the total sizes of tensors $\mathcal{M}^{\alpha}_t$ and $\mathcal{H}_{t-1}$, respectively, then
\begin{align}
\left(\frac{\partial \mathcal{M}^{\alpha}_t}{\partial \mathcal{H}_{t-1}}\right)_{|\mathcal{M}| \times |\mathcal{H}|} = \mathbf W_{\alpha D}\otimes \cdots \otimes \mathbf W_{\alpha 1}, \label{Eq19}
\end{align}
where the matricized form has been applied and $\otimes$ means the Kronecker product of matrices. And
\begin{align}
\frac{\partial \mathcal{M}^{\alpha}_{t(d)}}{\partial \mathbf W_{\alpha d}} = [(&\mathbf W_{\alpha D} \otimes \cdots \otimes \mathbf W_{\alpha (d-1)}\otimes \mathbf W_{\alpha (d+1)} \notag\\
& \otimes \cdots \otimes \mathbf W_{\alpha 1}) \mathcal{H}^T_{(t-1)(d)}]\otimes \mathbf I_{h_d\times h_d} \label{Eq20}\\
\frac{\partial \mathcal{M}^{\alpha}_{t(d)}}{\partial \mathbf U_{\alpha d}} = [(&\mathbf U_{\alpha D} \otimes \cdots \otimes \mathbf U_{\alpha (d-1)}\otimes \mathbf U_{\alpha (d+1)} \notag\\
& \otimes \cdots \otimes \mathbf U_{\alpha 1})\mathcal{X}^T_{t(d)}]\otimes \mathbf I_{x_d\times x_d} \label{Eq21}\\
\left(\frac{\partial \mathcal{M}^{\alpha}_t}{\partial \mathcal{B}_{\alpha}}\right)&_{|\mathcal{M}| \times |\mathcal{M}|} = \mathbf I_{|\mathcal{M}|\times |\mathcal{M}|} \label{Eq22}
\end{align}
where the subscript $(d)$ means the $d$-mode matricization of a tensor, and both $h_d$ and $x_d$ are the dimension of mode $d$ of tensors $\mathcal{H}_{t-1}$ and $\mathcal{X}_t$, respectively.
\end{lemma}

\begin{figure*}[t]
\centering
\includegraphics[width=13cm]{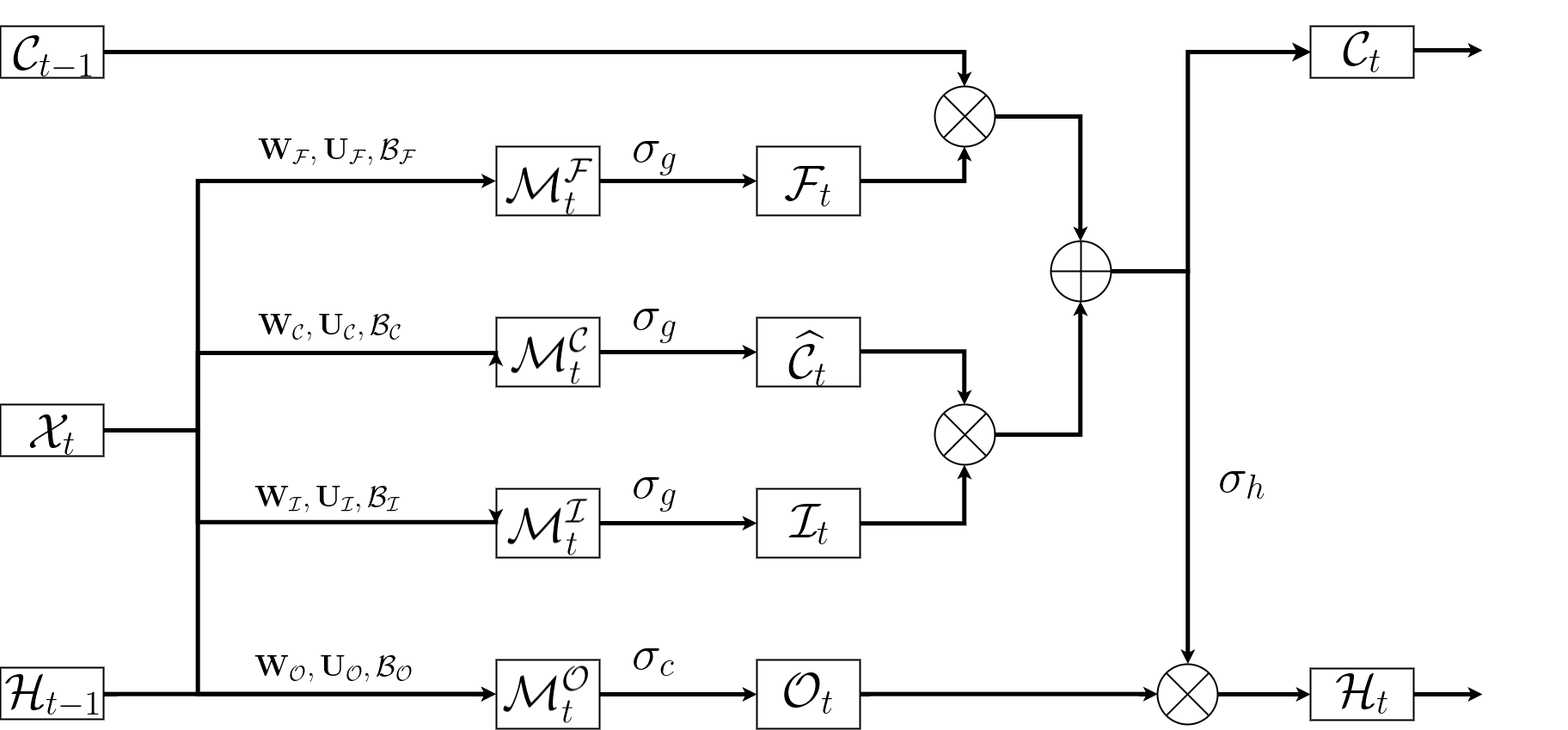}
\caption{LSTM Computation Flow. 
}
\label{Figure1}
\end{figure*}

First let us derive the BP algorithm for each LSTM step. The computation flow defined by equations \eqref{Eq3} - \eqref{Eq8} can be shown in Fig.~\ref{Figure1}. Along the time, $\mathcal{C}_t$ will be forwarded to the next LSTM step, while the hidden tensorial $\mathcal{H}_t$ will be carried onto the next step and also output an extra layer to match the response $\mathcal{Y}_t$ at $t$. Hence in the BP algorithm on LSTM unit, there could be two pieces of information, one from the next LSTM unit, denoted by $\frac{\partial l}{\partial \mathcal{H}_t}$, and the other from the output loss $l_t$, denoted by  $\frac{\partial l_o}{\partial \mathcal{H}_t}$. Thus the combined derivative information to be further backpropagated is
\begin{align}
\frac{\partial l}{\partial_h \mathcal{H}_t} =  \frac{\partial l}{\partial \mathcal{H}_t} + \frac{\partial l_o}{\partial \mathcal{H}_t}. \label{Eq23}
\end{align}

We know that $\frac{\partial l_o}{\partial \mathcal{H}_t} = 0$ if there is no response $\mathcal{Y}_t$ at time $t$.

According to the flows from $\mathcal{C}_{t-1}$, $\mathcal{F}_t$, $\widehat{\mathcal{C}}_{t}$, $\mathcal{I}_t$ and $\mathcal{O}_t$, respectively, to both $\mathcal{C}_t$ and $\mathcal{H}_t$ and the chain rule, it is easy to read from the diagram
\begin{align}
\frac{\partial l}{\partial \mathcal{C}_{t-1}} = & \left(\frac{\partial l}{\partial \mathcal{C}_{t}} + \sigma '_h (\mathcal{C}_t) \circ \frac{\partial l}{\partial_h\mathcal{H}_t} \right)\circ \mathcal{F}_t \label{Eq24}\\
\frac{\partial l}{\partial \mathcal{F}_{t}} = & \left(\frac{\partial l}{\partial \mathcal{C}_{t}} + \sigma '_h (\mathcal{C}_t) \circ \frac{\partial l}{\partial_h\mathcal{H}_t}\right) \circ \mathcal{C}_{t-1} \label{Eq25}\\
\frac{\partial l}{\partial \widehat{\mathcal{C}}_{t}} = & \left(\frac{\partial l}{\partial \mathcal{C}_{t}} + \sigma '_h (\mathcal{C}_t) \circ \frac{\partial l}{\partial_h\mathcal{H}_t} \right)\circ \mathcal{I}_{t} \label{Eq26}\\
\frac{\partial l}{\partial \mathcal{I}_{t}} = & \left(\frac{\partial l}{\partial \mathcal{C}_{t}} + \sigma '_h (\mathcal{C}_t) \circ \frac{\partial l}{\partial_h\mathcal{H}_t} \right) \circ \widehat{\mathcal{C}}_{t}\label{Eq27}\\
\frac{\partial l}{\partial \mathcal{O}_{t}} = &   \sigma_h (\mathcal{C}_t) \circ \frac{\partial l}{\partial_h\mathcal{H}_t} \label{Eq28}
\end{align}

Let us introduce two operators: for any tensor $\mathcal{M}$, denote the vectorization as $\mathbf m = \text{Vec}(\mathcal{M})$ while its inverse operator $\text{iVec}(\mathbf m) = \mathcal{M}$. Then we will have
\begin{align}
\frac{\partial l}{\partial \mathcal{H}_{t-1}} = & \text{iVec}\left( \text{Vec}(\frac{\partial l}{\partial \mathcal{F}_{t}} \circ \sigma '_g(\mathcal{M}^f_t))^T \left(\frac{\partial \mathcal{M}^{f}_t}{\partial \mathcal{H}_{t-1}}\right)\right) \notag\\
+& \text{iVec}\left( \text{Vec}(\frac{\partial l}{\partial \widehat{\mathcal{C}}_{t}} \circ \sigma '_g(\mathcal{M}^c_t))^T \left(\frac{\partial \mathcal{M}^{c}_t}{\partial \mathcal{H}_{t-1}}\right)\right) \notag\\
+&\text{iVec}\left( \text{Vec}(\frac{\partial l}{\partial \mathcal{I}_{t}} \circ \sigma '_g(\mathcal{M}^i_t))^T \left(\frac{\partial \mathcal{M}^{i}_t}{\partial \mathcal{H}_{t-1}}\right)\right) \notag\\
+&\text{iVec}\left( \text{Vec}(\frac{\partial l}{\partial \mathcal{O}_{t}} \circ \sigma '_g(\mathcal{M}^o_t))^T \left(\frac{\partial \mathcal{M}^{o}_t}{\partial \mathcal{H}_{t-1}}\right)\right). \label{Eq29}
\end{align}

Similarly we have, at time $t$,
\begin{align}
\left.\frac{\partial l}{\partial \mathbf W_{\alpha d}}\right|_{t} =& \text{iVec}\left(\text{Vec}(\frac{\partial l}{\partial \mathcal{A}_t} \circ \sigma '_g(\mathcal{M}^{\alpha}_t))^T \frac{\partial \mathcal{M}^{\alpha}_{t(d)}}{\partial \mathbf W_{\alpha d}} \right) \label{Eq30}\\
\left.\frac{\partial l}{\partial \mathbf U_{\alpha d}}\right|_{t} =& \text{iVec}\left(\text{Vec}(\frac{\partial l}{\partial \mathcal{A}_t} \circ \sigma '_g(\mathcal{M}^{\alpha}_t))^T \frac{\partial \mathcal{M}^{\alpha}_{t(d)}}{\partial \mathbf U_{\alpha d}} \right) \label{Eq31}\\
\left.\frac{\partial l}{\partial \mathcal{B}_{\alpha}}\right|_{t} =& \left[\text{sum}\left(\frac{\partial l}{\partial \mathcal{A}_t} \circ \sigma '_g(\mathcal{M}^{\alpha}_t)\right)\right] \,(\text{const tensor}) \label{Eq32}
\end{align}
where $(\mathcal{A}, \alpha) = (\mathcal{F}, f)$ or $(\widehat{\mathcal{C}},c)$ or $(\mathcal{I}, i)$ or $(\mathcal{O}, o)$.

Finally the overall derivatives for the parameters are given by 
\begin{align}
 \frac{\partial l}{\partial \mathbf W_{\alpha d}} =& \sum^{T}_{t=1}\left.\frac{\partial l}{\partial \mathbf W_{\alpha d}}\right|_{t} \label{Eq33}\\
  \frac{\partial l}{\partial \mathbf U_{\alpha d}} =& \sum^{T}_{t=1}\left.\frac{\partial l}{\partial \mathbf U_{\alpha d}}\right|_{t} \label{Eq34}\\
   \frac{\partial l}{\partial \mathcal{B}_{\alpha}} =& \sum^{T}_{t=1}\left.\frac{\partial l}{\partial \mathcal{B}_{\alpha}}\right|_{t}. \label{Eq35}
\end{align}

Let us consider loss functions defined in \eqref{Eq15} and \eqref{Eq16}. First, we note that $\frac{\partial \ell}{\partial \mathcal{H}_T}=0$. In the case of \eqref{Eq16},   $\frac{\partial \ell_o}{\partial \mathcal{H}_t}$ is calculated according to the given layer and cost function for the response $\mathcal{Y}_t$ for $t=1, 2, ..., T$. In the case of \eqref{Eq15}, we only have the information $\frac{\partial \ell_o}{\partial \mathcal{H}_t}$ when $t=T$ otherwise 0. Hence \eqref{Eq23} will be calculated accordingly.  

We summarize the derivative BP algorithm for LSTM in Algorithm \ref{Alg1}
\begin{algorithm}
\renewcommand{\algorithmicrequire}{\textbf{Input:}}
\renewcommand\algorithmicensure {\textbf{Output:} } 
\caption{The Derivative BP Algorithm for LSTM (for a single training data)} \label{Alg1}
\begin{algorithmic}[1]
  \REQUIRE The training case $\{(\mathcal{X}_{j1}, \mathcal{X}_{j2}, \cdots, \mathcal{X}_{jT}), \mathcal{Y}_{jT}\}$ (or $\{(\mathcal{X}_{j1}, \mathcal{X}_{j2}, \cdots, \mathcal{X}_{jT}), (\mathcal{Y}_{j1}, \mathcal{Y}_{j2}, \cdots, \mathcal{Y}_{jT})\}$)  
  \ENSURE $\frac{\partial l}{\partial \mathbf W_{\alpha d}}$, $\frac{\partial l}{\partial \mathbf U_{\alpha d}}$, and $\frac{\partial l}{\partial \mathcal{B}_{\alpha}}$ for $\alpha = f, c, i, o$ and $d=1, 2, ..., D$.
\STATE Set $\frac{\partial \ell}{\partial \mathcal{C}_T}=0$, $\frac{\partial \ell}{\partial \mathcal{H}_T}=0$ and calculate  $\frac{\partial \ell_o}{\partial \mathcal{H}_T}$ according to the output layer for $\mathcal{Y}_{jT}$ at time $T$; 
 \FOR {$t=T$ to $1$}
 \STATE Use \eqref{Eq23} to calculate $\frac{\partial l}{\partial_h\mathcal{H}_t}$ while, for $t<T$, $\frac{\partial \ell_o}{\partial \mathcal{H}_t} = 0$ if there is no response at $t$; otherwise it is calculated according to the output loss at $t$;
    \STATE Use \eqref{Eq25} - \eqref{Eq28} to calculate  $\frac{\partial \ell}{\partial \mathcal{F}_{t}}$,  $\frac{\partial \ell}{\partial \widehat{\mathcal{C}}_{t}}$, $\frac{\partial \ell}{\partial \mathcal{I}_{t}}$ and $\frac{\partial \ell}{\partial \mathcal{O}_{t}}$;
\STATE Use \eqref{Eq20} - \eqref{Eq22} to calculate $\frac{\partial \mathcal{M}^{\alpha}_{t(d)}}{\partial \mathbf W_{\alpha d}}$ and $\frac{\partial \mathcal{M}^{\alpha}_{t(d)}}{\partial \mathbf U_{\alpha d}}$;   
\STATE Use \eqref{Eq30} - \eqref{Eq32} to calculate $\left.\frac{\partial l}{\partial \mathbf W_{\alpha d}}\right|_{t}$, $\left.\frac{\partial l}{\partial \mathbf U_{\alpha d}}\right|_{t}$ and $\left.\frac{\partial l}{\partial \mathcal{B}_{\alpha }}\right|_{t}$
\STATE Use \eqref{Eq19} to calculate  $\frac{\partial \mathcal{M}^{\alpha}_t}{\partial \mathcal{H}_{t-1}}$ for $\alpha = f, c, i, o$;  
\STATE Use \eqref{Eq24} to update $\frac{\partial \ell}{\partial \mathcal{C}_{t-1}}$ and \eqref{Eq29} for $\frac{\partial \ell}{\partial \mathcal{H}_{t-1}}$
 \ENDFOR
\STATE Use \eqref{Eq33} - \eqref{Eq35} to summarize $\frac{\partial l}{\partial \mathbf W_{\alpha d}}$, $\frac{\partial l}{\partial \mathbf U_{\alpha d}}$, and $\frac{\partial l}{\partial \mathcal{B}_{\alpha}}$ for outputs.
\end{algorithmic}
\end{algorithm}

\begin{figure*}[!h]
\centering
\includegraphics[width=13cm]{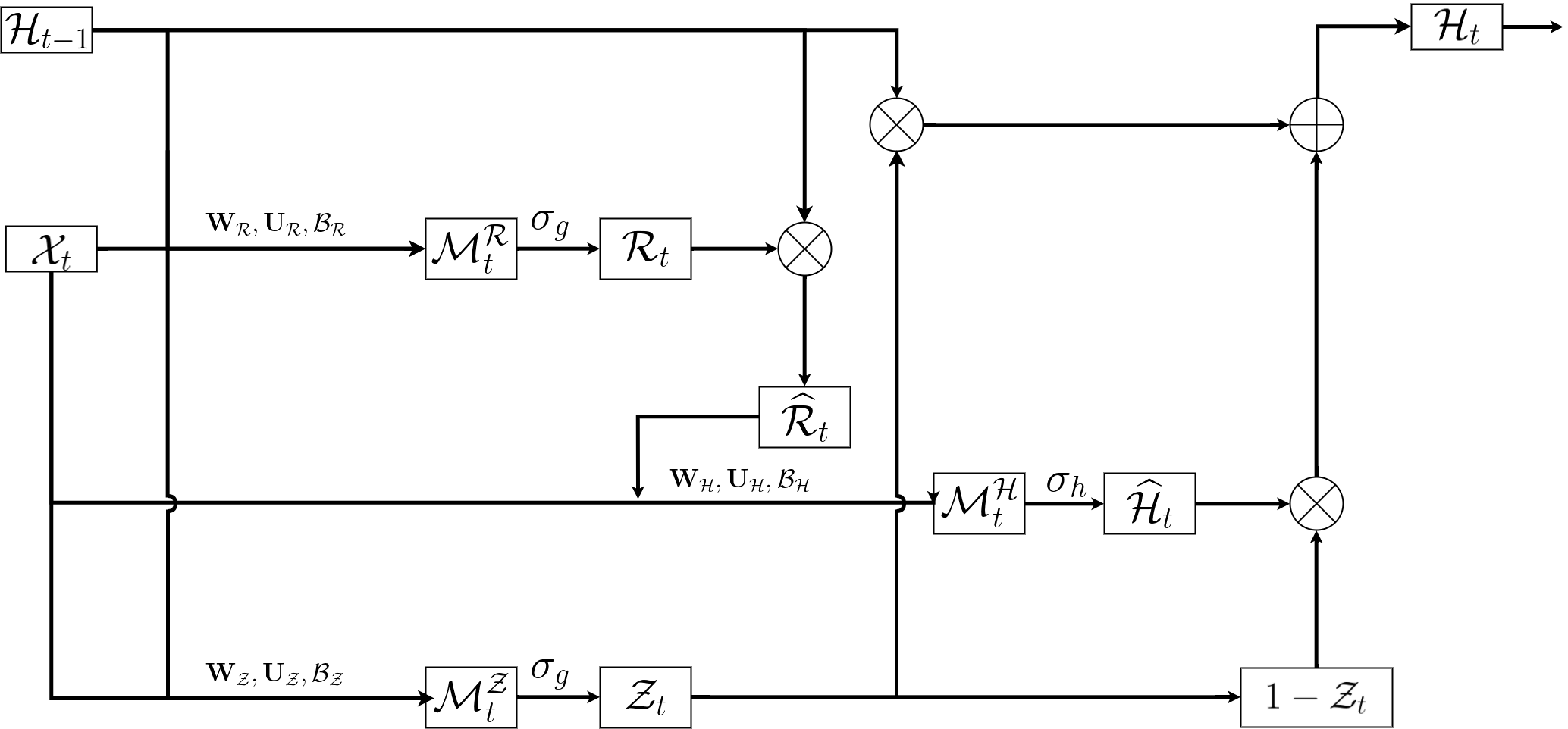}
\caption{GRU Computation Flow. 
}
\label{Figure2}
\end{figure*}

The BP algorithm for GRU can be derived in a similar way by looking at the  computation flow defined by equations \eqref{Eq9} - \eqref{Eq13}, as shown in Fig.~\ref{Figure2}

Similarly, the backpropagated information from time $t+1$ is $\frac{\partial \ell}{\partial_h \widehat{\mathcal{H}}_t}$ as in \eqref{Eq23}. Hence 
\begin{align}
\frac{\partial \ell}{\partial \widehat{\mathcal{H}}_t} =& ( 1-\mathcal{Z}_t)\circ \frac{\partial \ell}{\partial_h {\mathcal{H}}_t} \label{Eq36}\\
\frac{\partial \ell}{\partial {\mathcal{Z}}_t} =& (\mathcal{H}_{t-1} -\widehat{\mathcal{H}}_t)\circ \frac{\partial \ell}{\partial_h {\mathcal{H}}_t} \label{Eq37}\\
\frac{\partial \ell}{\partial {\mathcal{R}}_t} = & \mathcal{H}_{t-1}\circ \text{iVec}\left( \text{Vec}(\frac{\partial l}{\partial \widehat{\mathcal{H}}_{t}} \circ \sigma '_h(\mathcal{M}^h_t))^T \left(\frac{\partial \mathcal{M}^{h}_t}{\partial \widehat{\mathcal{R}}_{t}}\right)\right)
\label{Eq38}
\end{align}
\begin{align}
\frac{\partial \ell}{\partial {\mathcal{H}}_{t-1}} =& \text{iVec}\left( \text{Vec}(\frac{\partial l}{\partial {\mathcal{R}}_{t}} \circ \sigma '_g(\mathcal{M}^r_t))^T \left(\frac{\partial \mathcal{M}^{r}_t}{\partial {\mathcal{R}}_{t}}\right)\right) \notag\\
&+\text{iVec}\left( \text{Vec}(\frac{\partial l}{\partial {\mathcal{Z}}_{t}} \circ \sigma '_g(\mathcal{M}^z_t))^T \left(\frac{\partial \mathcal{M}^{z}_t}{\partial {\mathcal{Z}}_{t}}\right)\right) \notag\\
& + \mathcal{Z}_t \circ \frac{\partial \ell}{\partial_h  {\mathcal{H}}_t} \label{Eq39}
\end{align}

The derivatives with respect to all three sets of parameters can be obtained from \eqref{Eq30} - \eqref{Eq32} with $(\mathcal{A}, \alpha) = (\mathcal{R}, r)$, $(\mathcal{Z}, z)$ and $(\widehat{\mathcal{H}}, h)$.

The derivative BP algorithm for GRU is summarized in Algorithm \ref{Alg2}
\begin{algorithm}
\renewcommand{\algorithmicrequire}{\textbf{Input:}}
\renewcommand\algorithmicensure {\textbf{Output:} } 
\caption{The Derivative BP Algorithm for GRU (for a single training data)} \label{Alg2}
\begin{algorithmic}[1]
  \REQUIRE The training case $\{(\mathcal{X}_{j1}, \mathcal{X}_{j2}, \cdots, \mathcal{X}_{jT}), \mathcal{Y}_{jT}\}$ (or $\{(\mathcal{X}_{j1}, \mathcal{X}_{j2}, \cdots, \mathcal{X}_{jT}), (\mathcal{Y}_{j1}, \mathcal{Y}_{j2}, \cdots, \mathcal{Y}_{jT})\}$)  
  \ENSURE $\frac{\partial l}{\partial \mathbf W_{\alpha d}}$, $\frac{\partial l}{\partial \mathbf U_{\alpha d}}$, and $\frac{\partial l}{\partial \mathcal{B}_{\alpha}}$ for $\alpha = r, z, h$ and $d=1, 2, ..., D$.
\STATE Set $\frac{\partial \ell}{\partial \mathcal{H}_T}=0$ and calculate  $\frac{\partial \ell_o}{\partial \mathcal{H}_T}$ according to the output layer for $\mathcal{Y}_{jT}$ at time $T$; 
 \FOR {$t=T$ to $1$}
 \STATE Use \eqref{Eq23} to calculate $\frac{\partial l}{\partial_h\mathcal{H}_t}$ while, for $t<T$, $\frac{\partial \ell_o}{\partial \mathcal{H}_t} = 0$ if there is no response at $t$; otherwise it is calculated according to the output loss at $t$;
 \STATE Use \eqref{Eq19} to calculate  $\frac{\partial \mathcal{M}^{\alpha}_t}{\partial \mathcal{H}_{t-1}}$, for $\alpha = r, z$, and $\frac{\partial \mathcal{M}^{h}_t}{\partial \widehat{\mathcal{R}}_{t}}$;  
    \STATE Use \eqref{Eq36} - \eqref{Eq38} to calculate  $\frac{\partial \ell}{\partial \widehat{\mathcal{H}}_{t}}$,  $\frac{\partial \ell}{\partial {\mathcal{Z}}_{t}}$ and $\frac{\partial \ell}{\partial \mathcal{R}_{t}}$;
\STATE Use \eqref{Eq20} - \eqref{Eq22} to calculate $\frac{\partial \mathcal{M}^{\alpha}_{t(d)}}{\partial \mathbf W_{\alpha d}}$ and $\frac{\partial \mathcal{M}^{\alpha}_{t(d)}}{\partial \mathbf U_{\alpha d}}$;   
\STATE Use \eqref{Eq30} - \eqref{Eq32} to calculate $\left.\frac{\partial l}{\partial \mathbf W_{\alpha d}}\right|_{t}$, $\left.\frac{\partial l}{\partial \mathbf U_{\alpha d}}\right|_{t}$ and $\left.\frac{\partial l}{\partial \mathcal{B}_{\alpha }}\right|_{t}$

\STATE Use  \eqref{Eq39} to update $\frac{\partial \ell}{\partial \mathcal{H}_{t-1}}$
 \ENDFOR
\STATE Use \eqref{Eq33} - \eqref{Eq35} to summarize $\frac{\partial l}{\partial \mathbf W_{\alpha d}}$, $\frac{\partial l}{\partial \mathbf U_{\alpha d}}$, and $\frac{\partial l}{\partial \mathcal{B}_{\alpha}}$ for outputs.
\end{algorithmic}
\end{algorithm}

\section{Experiments} \label{Sec:4} 
\subsection{Data Description}
To assess the performance of tRNN on the real world data, we conduct an empirical study. In this study, the data is collected from Integrated Crisis Warning System (ICEWS) \footnote{\url{http://www.lockheedmartin.com/us/products/W-ICEWS/iData.html}} which is the same weekly dataset applied in the study of MLTR \cite{Hoff2015} for the relationship between 25 countries in four types of actions:  material cooperation, material conflict, verbal cooperation and verbal conflict, from 2004 to mid-2014. Thus at any particular time point, the data is a 3D tensor of dimensions $25\times 25\times 4$. That is, each input ${\mathcal{X}}_t \in \mathbb{R}^{25\times 25\times 4}$. To explore different types of patterns often seen in relational data and social networks, we organise explanatory tensors $\mathcal{X}$ in the following different ways.  

As done in \cite{Hoff2015} we construct the target tensor $\mathcal{Y}_t$ at time $t$ as the lagged $\mathcal{X}_{t-1}$. In total, we construct an overall dataset $\{(\mathcal{X}_t, \mathcal{Y}_t)\}^{543}_{t=1}$ of size 543 in which all $\mathcal{X}_t$ and $\mathcal{Y}_t$ are 3D tensors. Further we take $T=7$ as the period of time series sections, and use 90\% data for training, as defined in the following two cases:

\textit{Case I:} We use LSTM in terms of a many-to-one recurrent model  with the training dataset as follows, 
\begin{align*}
\mathcal{D}_1 = \{(\mathcal{X}_{j},\mathcal{X}_{j+1}, ..., \mathcal{X}_{j+6}), \mathcal{Y}_{j+6}\}^{488}_{j=1}
\end{align*}
and the remaining 55 data will be used for testing. 

\textit{Case II:} We use LSTM in terms of many-to-many recurrent model   with the training dataset as follows,
\begin{align*}
\mathcal{D}_2 = \{(\mathcal{X}_{j},\mathcal{X}_{j+1}, ..., \mathcal{X}_{j+6}), (\mathcal{Y}_{j}, \mathcal{Y}_{j+1}, ...,\mathcal{Y}_{j+6})\}^{488}_{j=1}
\end{align*}
and the remaining 55 data will be used for testing. 

\subsection{Experiment Setting and Results}
Our intention in these two experiments is to quickly 
demonstrate how tLSTM works with the most possible convenience. tGRU can give similar results. 

In Case I, we set the size of the hidden nodes to be $50 \times 50$, doubled the input size $25\times 25$.  In Case II, we use half the input size for the hidden nodes, i.e., $14\times 14$. We also use a regulariser for all the matrix coefficients $\mathbf W$ and $\mathcal U$, i.e., adding the following to the loss function to have a regularised objective function,
\[
\lambda \left( \sum_{\text{all }\mathbf W} \|\mathbf W\|^2_F + \sum_{\text{all }\mathbf U} \|\mathbf U\|^2_F\right)
\]
where $\|\cdot\|_F$ is the Frobenius norm for matrices and $\lambda>0$ is a parameter to trade-off between the loss and the regulariser. In our experiment we empirically set $\lambda = 0.01$.  This parameter should be optimised by using a set of valid dataset.
 
\begin{figure*}[h]
\centering
\subfigure[]{\includegraphics[width=6.5cm]{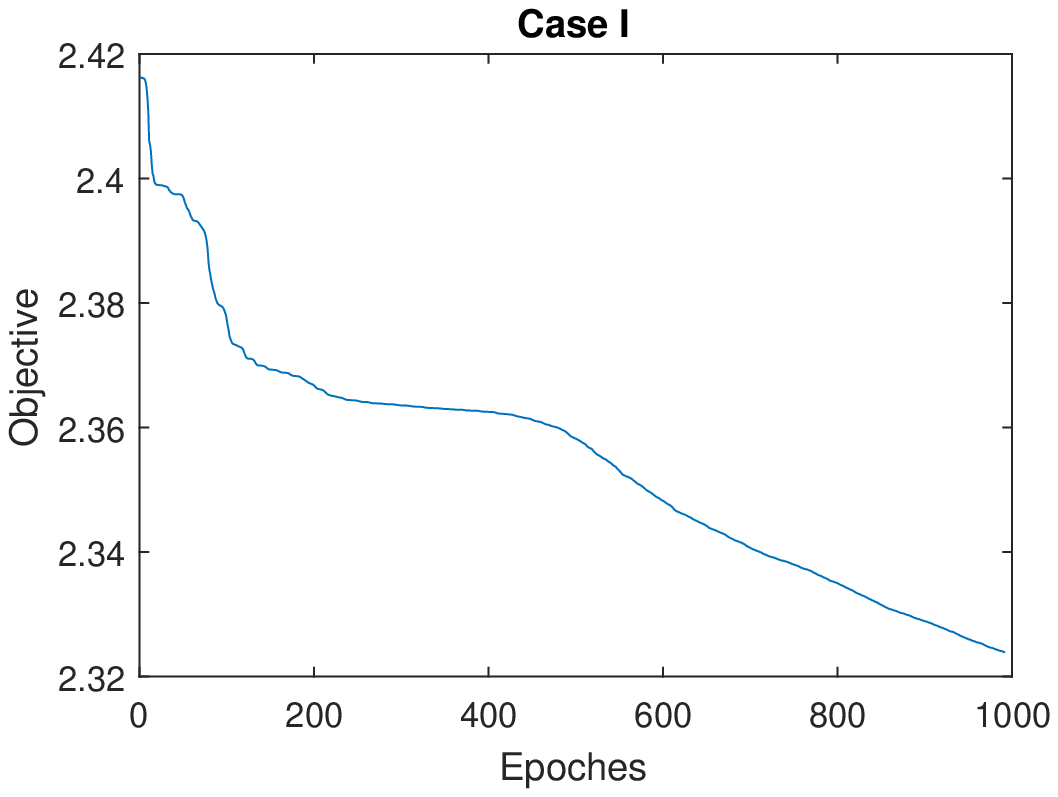}}\ \ \ \ 
\subfigure[]{\includegraphics[width=6.3cm]{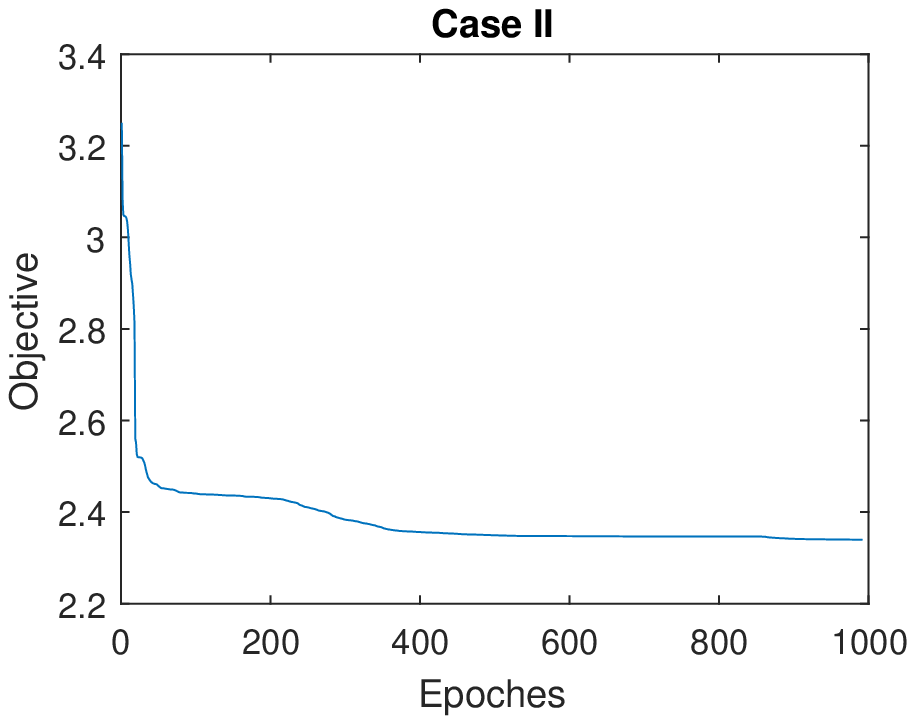}}
\caption{Convergence Trends for Two Cases.}
\label{Figure3}
\end{figure*}

Fig. \ref{Figure3} shows the convergence trends for both cases in training process with 1000 epoches. The test errors are 0.0081 for Case I and 0.0082 for Case II.
\section{Conclusion} \label{Sec:5} 
In this paper, we introduced the new recurrent neural networks  for
high-order tensor data. Two special recurrent structures, i.e., tLSTM and tGRU, are proposed with detailed BP algorithm derivation. Two simple experiments have demonstrated the performance of the new recurrent neural networks. More experiments shall be conducted to demonstrate its efficiency and accuracy against the existing neural networks. We also intend to explore more applications such as for video data analysis. 

\bibliographystyle{IEEEtran}
\bibliography{newRef}

\begin{thebibliography}{10}
\providecommand{\url}[1]{#1}
\csname url@samestyle\endcsname
\providecommand{\newblock}{\relax}
\providecommand{\bibinfo}[2]{#2}
\providecommand{\BIBentrySTDinterwordspacing}{\spaceskip=0pt\relax}
\providecommand{\BIBentryALTinterwordstretchfactor}{4}
\providecommand{\BIBentryALTinterwordspacing}{\spaceskip=\fontdimen2\font plus
\BIBentryALTinterwordstretchfactor\fontdimen3\font minus
  \fontdimen4\font\relax}
\providecommand{\BIBforeignlanguage}[2]{{%
\expandafter\ifx\csname l@#1\endcsname\relax
\typeout{** WARNING: IEEEtran.bst: No hyphenation pattern has been}%
\typeout{** loaded for the language `#1'. Using the pattern for}%
\typeout{** the default language instead.}%
\else
\language=\csname l@#1\endcsname
\fi
#2}}
\providecommand{\BIBdecl}{\relax}
\BIBdecl

\bibitem{GaoGuoWang2017}
J.~Gao, Y.~Guo, and Z.~Wang, ``Matrix neural networks,'' in \emph{Proceedings
  of the14th International Symposium on Neural Networks (ISNN)}, vol. Accepted,
  Sapporo, Japan, 2017, pp. 1--10.

\bibitem{IonescuVantzosSminchisescu2015a}
C.~Ionescu, O.~Vantzos, and C.~Sminchisescu, ``Matrix backpropagation for deep
  networks with structured layers,'' in \emph{Proceedings of the IEEE
  International Conference on Computer Vision (ICCV)}, 2015, pp. 2965--2973.

\bibitem{BaiZhangGao2017}
M.~Bai, B.~Zhang, and J.~Gao, ``Tensorial neural networks and its application
  in longitudinal network data analysis,'' in \emph{submitted to the 24th
  International Conference On Neural Information Processing (ICONIP)}, 2017.

\bibitem{ChienBao2017}
\BIBentryALTinterwordspacing
J.-T. Chien and Y.-T. Bao, ``Tensor-factorized neural networks,'' \emph{IEEE
  Transactions on Neural Networks and Learning Systems}, vol.~PP, 2017.
  [Online]. Available: \url{http://ieeexplore.ieee.org/document/7902201/}
\BIBentrySTDinterwordspacing

\bibitem{LiTarlowBrockschmidtZemel2016}
Y.~Li, D.~Tarlow, M.~Brockschmidt, and R.~Zemel, ``Gated graph sequence neural
  networks,'' in \emph{Proceedings of International Conference on Learning
  Representations (ICLR)}, 2016.

\bibitem{SeoDefferrardVandergheynstBresson2017}
Y.~Seo, M.~Defferrard, P.~Vandergheynst, and X.~Bresson, ``Structured sequence
  modeling with graph convolutional recurrent networks,'' in \emph{Proceedings
  of International Conference on Learning Representation}, 2017.

\bibitem{HuangGool2017}
Z.~Huang and L.~V. Gool, ``A {R}iemannian network for spd matrix learning,'' in
  \emph{Proceedings of the Thirty-First AAAI Conference on Artificial
  Intelligence (AAAI-17)}, 2017.

\bibitem{MulderBethardMoens2015}
W.~D. Mulder, S.~Bethard, and M.-F. Moens, ``A survey on the application of
  recurrent neural networks to statistical language modeling,'' \emph{Computer
  Speech \& Language}, vol.~30, no.~1, pp. 61--98, 2015.

\bibitem{GravesMohamedHinton2013}
A.~Graves, A.~rahman Mohamed, and G.~Hinton, ``Speech recognition with deep
  recurrent neural networks,'' in \emph{Proceedings of IEEE International
  Conference on Acoustics, Speech and Signal Processing (ICASSP)}, 2013.

\bibitem{LiuShahroudyXuWang2016}
J.~Liu, A.~Shahroudy, D.~Xu, and G.~Wang, ``Spatio-temporal {LSTM} with trust
  gates for {3D} human action recognition,'' in \emph{Proceedings of the
  European Conference on Computer Vision (ECCV)}, 2016, pp. 816--833.

\bibitem{ZhangLanXingZengXueZheng2017}
P.~Zhang, C.~Lan, J.~Xing, W.~Zeng, J.~Xue, and N.~Zheng, ``View adaptive
  recurrent neural networks for high performance human action recognition from
  skeleton data,'' \emph{arXiv:1703.08274}, vol.~1, 2017.

\bibitem{GanyPuyHenaoyLiyHezCarin2016}
Z.~Gany, Y.~Puy, R.~Henaoy, C.~Liy, X.~Hez, and L.~Carin, ``Learning generic
  sentence representations using convolutional neural networks,''
  \emph{arXiv:1611.07897}, vol.~2, 2016.

\bibitem{JoseCisseFleuret2017}
C.~Jose, M.~Ciss\'{e}, and F.~Fleuret, ``Kronecker recurrent units,''
  \emph{arXiv:1705.10142}, 2017.

\bibitem{Elman1990}
J.~L. Elman, ``Finding structure in time,'' \emph{Cognitive Science}, vol.~14,
  no.~2, pp. 179--211, 1990.

\bibitem{Jordan1997}
M.~I. Jordan, \emph{Serial Order: A Parallel Distributed Processing Approach},
  ser. Advances in Psychology: Neural-Network Models of Cognition.\hskip 1em
  plus 0.5em minus 0.4em\relax Elsevier, 1997, vol. 121, ch.~25, pp. 471--495.

\bibitem{HochreiterSchmidhuber1997}
S.~Hochreiter and J.~Schmidhuber, ``Long short-term memory". .'' \emph{Neural
  Computation}, vol.~9, no.~8, pp. 1735--1780, 1997.

\bibitem{GersSchmidhuberCummins2000}
F.~A. Gers, J.~Schmidhuber, and F.~Cummins, ``Learning to forget: Continual
  prediction with {LSTM},'' \emph{Neural Computation}, vol.~12, no.~10, pp.
  2451--2471, 2000.

\bibitem{KoldaBader2009}
T.~G. Kolda and B.~W. Bader, ``Tensor decompositions and applications,''
  \emph{SIAM Review}, vol.~51, no.~3, pp. 455--500, 2009.

\bibitem{ChoMerrieenboerGuelccehreBahdanauBougaresSchwenkBengio2014}
\BIBentryALTinterwordspacing
K.~Cho, B.~van Merri{\"{e}}nboer, {\c C}.~G{\"{u}}l{\c c}ehre, D.~Bahdanau,
  F.~Bougares, H.~Schwenk, and Y.~Bengio, ``Learning phrase representations
  using rnn encoder--decoder for statistical machine translation,'' in
  \emph{Proceedings of the 2014 Conference on Empirical Methods in Natural
  Language Processing (EMNLP)}.\hskip 1em plus 0.5em minus 0.4em\relax Doha,
  Qatar: Association for Computational Linguistics, Oct. 2014, pp. 1724--1734.
  [Online]. Available: \url{http://www.aclweb.org/anthology/D14-1179}
\BIBentrySTDinterwordspacing

\bibitem{Kolda2006}
T.~G. Kolda, ``{Multilinear Operators for Higher-Order Decompositions},''
  Sandia National Laboratories, Technical report, 2006.

\bibitem{Hoff2015}
P.~D. Hoff, ``{Multilinear Tensor Regression for Longitudinal Relational
  Data},'' \emph{Ann. Appl. Stat.}, vol.~9, no.~3, pp. 1169--1193, 2015.

\end{thebibliography}
\end{document}